\documentclass[11pt]{article}
\usepackage{coling2020}
\usepackage[utf8]{inputenc}
\usepackage[T2A,T1]{fontenc}
\usepackage[russian,english]{babel}
\usepackage{times}
\usepackage{url}
\usepackage{latexsym}
\usepackage{graphicx}
\usepackage{booktabs}
\usepackage{hyperref}
\usepackage{wrapfig}
\usepackage{relsize}
\usepackage{multirow}

\newcommand{\sense}[1] {{\smaller[0.5]`\textsc{{#1}}'}}

\colingfinalcopy 

\title{RuSemShift: a dataset of historical lexical semantic change in Russian}

\author{Julia Rodina \\
  National Research University Higher School of Economics \\
  Moscow, Russia \\
  {\tt julia.rodina97@gmail.com} \\\And
  Andrey Kutuzov \\
  University of Oslo \\
  Oslo, Norway \\
  {\tt andreku@ifi.uio.no} \\}

\date{}

\begin{document}
\maketitle
\begin{abstract}
  We present RuSemShift, a large-scale manually annotated test set for the task of semantic change modeling in Russian for two long-term time period pairs: from the pre-Soviet through the Soviet times and from the Soviet through the post-Soviet times. Target words were annotated by multiple crowd-source workers. The annotation process was organized following the DURel framework and was based on sentence contexts extracted from the Russian National Corpus. Additionally, we report the performance of several distributional approaches on RuSemShift, achieving promising results, which at the same time leave room for other researchers to improve.
\end{abstract}

\section{Introduction}

Language is a constantly changing system by its nature, since it is a method of communication and a social instrument. As such, it should meet the needs of the speakers, and thus it adapts to changes in the society and the ever-changing  world. As a part of language, lexical meaning also evolves over time, with words undergoing diachronic (or temporal) semantic shifts \cite{traugott2001regularity}.

Tracing semantic change can be important either in itself, as a linguistic study, or for practical downstream applications, for example, in socio-linguistic research.  Manual analysis of such shifts is time-consuming and laborious even after the emergence of large representative corpora, since one needs to look through a lot of examples and lexicographic resources which often do not record current lexical changes in language due to limited resources. Thus, researchers are trying to model these processes using  advanced  computational  approaches often  based  on distributional semantics and dense word embeddings \cite{kutuzov:survey,tang_2018}. 

However, this is still mostly done for English: often simply because manually annotated test data is not available for other languages. Recently, consistently annotated lexical semantic change test sets for multiple languages started to appear; see, for example, \newcite{schlechtweg2020semeval}. In this paper, we continue this vein of work by presenting \textit{RuSemShift}. \textit{RuSemShift}\footnote{\url{https://github.com/juliarodina/RuSemShift}} is the first historical semantic change dataset for Russian annotated according to the DURel framework \cite{durel} using a large crowd-sourcing platform, instead of personal intuitions of individual researchers. It allows to evaluate semantic change detection systems by their ability to estimate the shifts which occurred to Russian words either after 1917 (the fall of the Russian Empire) or after 1990 (the fall of the Soviet Union). 

The rest of the paper is organized as follows: in Section~\ref{sec:background}, we put our research in the context of the related work. In Section~\ref{sec:data}, we present the employed corpora and the process of the dataset creation. Section~\ref{sec:annotation} describes the annotation itself. In Section~\ref{sec:evaluation}, we empirically evaluate several existing semantic change detection algorithms (based on static and contextualized embeddings) on \textit{RuSemShift} to check its sanity. Section~\ref{conclusion} summarizes our contributions and outlines future research.

\section{Related work}
\label{sec:background}

Works on language change from general linguistics like \newcite{traugott2001regularity} or \newcite{twentywords} as a rule contain only a small number of hand-picked examples. The \textit{DatSemShift} database \cite{zalizniak2018catalogue} features more than 4 000 semantic shifts across 800 languages. But it is focused on cognitive proximities between pairs of linguistic meanings (with a limited set of pre-defined senses): in this paradigm, a semantic shift is just a case of extended polysemy. The \textit{DatSemShift} database is extremely useful for identifying recurring cross-linguistic semantic shifts, but it is difficult to employ it for evaluation of unsupervised semantic change detection systems.

To our knowledge, the first Russian test set to evaluate lexical semantic change detection systems was created by \newcite{kutuzov-kuzmenko}. They used prior linguistic work to manually collect Russian words which changed their meaning from the pre-Soviet times through the Soviet times. \newcite{kutuzov-kuzmenko} also employed static word embedding models trained on the corresponding Russian diachronic corpora to detect semantic change for the words from the dataset (distinguishing changed words from stable ones). They concluded that Kendall’s $\tau$ \cite{kendall1948rank} and Jaccard similarity \cite{Jaccard:1901} between nearest neighbor lists worked best in this tasks. More recent work by \newcite{fomin} extended this research to more granular time bins (periods of 1 year): they analyzed sequential pairs of word embedding models trained on yearly corpora of Russian news from 2000 up to 2014. To evaluate their system, they created a test set which contained human judgements about how much the meaning of a word has shifted over the given years. Using this ground truth, \newcite{fomin} evaluated 5 algorithms for semantic change detection and provided solid foundation for future research in in this direction for Russian.

However, these datasets suffer from two serious issues. First, they are either produced by a single person \cite{kutuzov-kuzmenko} or annotated by the paper authors themselves \cite{fomin}. This makes them inherently subjective. Second, each of this datasets is created and annotated in a different manner, making them inconsistent and not comparable nor to themselves neither to similar efforts made for other languages. In addition, the test set from \cite{kutuzov-kuzmenko} features only binary labels (shifted or stable), making it useless for research in graded semantic change detection.

The \textit{RuSemShift} test set we present in this paper makes partial use of these existing test sets, but all the words are re-annotated from scratch. In \textit{RuSemShift}, we adhere to the language-agnostic \textit{Diachronic Usage Relatedness} (DURel) semantic change annotation methodology  proposed in \newcite{durel}. The intended use of the DURel framework is the creation of consistently and robustly annotated test sets containing words labeled with the degrees of their diachronic lexical semantic change. \newcite{durel} presented a test set for German annotated this way. In addition, English, German, Latin and Swedish test sets used in the SemEval-2020 shared task on unsupervised lexical semantic change detection \cite{schlechtweg2020semeval} were also created following the same approach. Thus, we deem  DURel to be the current de-facto standard for the annotation of semantic change datasets. 

\begin{wrapfigure}{r}{0.4\textwidth}
\centering
\includegraphics[width=0.4\textwidth]{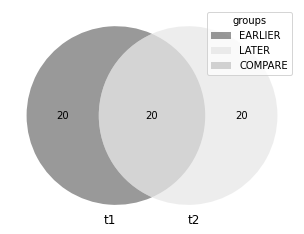}
\caption{Number of \textit{sentence pairs} for each word from the time period $t1$ (EARLIER), the time period $t2$ (LATER) and from both periods (COMPARE).}
\label{fig:Groups}
\end{wrapfigure}
DURel employs the notion of usage relatedness borrowed from research  on word sense disambiguation task \cite{brown}. The idea is to measure the degree of semantic change as a function of mean relatedness across pairs of word’s occurrences in different time periods. The annotators are not required to know anything about diachronic change: they are presented with two sentences (both containing a target word) and asked to estimate how similar are the senses in which the target word is used. They should choose a score from the 4-point scale described below in Section~\ref{sec:annotation}.

A sample of target word usage pairs from the first time period $t_1$ form the so called EARLIER group and the usage pairs from later time period $t_2$ form the LATER group. To cover the cases that cannot be tracked by comparing mean relatedness of two time periods, \newcite{durel} directly compare the `old' and `new' meanings by creating the additional COMPARE group which contains pairs where the first sentences is from the $t_1$ time period and the second sentence is from the $t_2$ time period (see Figure~\ref{fig:Groups}). Each group contains 20 sentence pairs randomly sampled from the corresponding corpora. Target word’s \textit{w} degree of semantic change is quantified with two measures: 

\begin{enumerate}
    \item $\Delta$LATER, which is the difference between mean relatedness of the EARLIER group and the LATER group: 
    
    $\Delta LATER(w) = Mean_{LATER}(w) - Mean_{EARLIER}(w)$;
    \item COMPARE, which is the mean relatedness within the COMPARE group: 
    
    $COMPARE(w) = Mean_{COMPARE}(w)$.
\end{enumerate}

Due to space limitations, it is impossible to provide a description of the advantages and disadvantages of these two measures here. We refer the reader to \newcite{durel} for the detailed discussion.

\section{Data Sources} \label{sec:data}
In this section, we describe our data sources. It can be looked at a sort of data statement \cite{bender-friedman-2018-data} for \textit{RuSemShift}. We first present the corpus we employ and its historical sub-corpora, and then move on to the process of preliminary target word selection. The data statement continues in the next Section~\ref{sec:annotation} with the description of our annotators and the annotation workflow in general.

\subsection{Corpora}
\textit{RuSemShift} covers 3 time periods of Russian language history, following \newcite{kutuzov-kuzmenko}:

\begin{enumerate}
  \item Texts produced from 1682 to 1916: the period of Russian monarchy before the revolution of 1917; further dubbed \textbf{pre-Soviet}.
  \item Texts produced from 1918 to 1990: the period of the existence of the Soviet Union; further dubbed \textbf{Soviet}.
  \item Texts produced from 1991 to 2017: the period after the fall of the Soviet Union: further dubbed \textbf{post-Soviet}.
\end{enumerate}

The wide coverage of these periods (several decades or even centuries each) makes it more likely that shifts in word usage will be caused by linguistic factors, and not only by extra-linguistic events. However, it is arguably still strongly influenced by cultural factors. The boundaries between time bins are related to rises and falls of political regimes, and some semantic shifts are inevitably related to that. Note that the first time period (the pre-Soviet times) is substantially longer temporally than the other two. This in theory can lead to multiple meaning shifts occurring \textit{within} this period. But this epoch is already the smallest in terms of corpus size (see below), so it was impossible to further divide it. 

Our annotation framework requires diachronic corpora with Russian texts created in the time periods listed above. As a source of such texts, we used the \textit{Russian National Corpus} (RNC).\footnote{\url{http://ruscorpora.ru}}
RNC is well balanced and contains Russian texts of diverse genres produced from the middle of the 18th century up to the beginning of the 21st century. Note that the corpus itself is not our contribution: it is a prior work.

The main RNC corpus size is about 320 million word tokens (including punctuation). Since all the texts are annotated with the date of their creation, it is straightforward to separate the corpus into time-specific sub-corpora:

\begin{itemize}
  \item 94 million tokens in the pre-Soviet sub-corpus;
  \item 123 million tokens in the Soviet sub-corpus;
  \item 107 million tokens in the post-Soviet sub-corpus.
\end{itemize}

\subsection{Preliminary word lists}

To construct word lists for further annotation, we handpicked words that presumably have undergone semantic changes in the Soviet period compared to the pre-Soviet period or in the post-Soviet compared to the Soviet period. For each of these word lists we also randomly sampled a set of `filler' or `distractor' words, trying to reproduce the part of speech and frequency percentile distributions of the original target words as closely as possible. The purpose of fillers (which are assumed to be semantically stable) is to be able to evaluate the performance of semantic change detection systems in a more realistic setup: they are supposed to predict low change scores for fillers (or no change at all).

\subsubsection{Pre-Soviet to Soviet dataset ($RuSemShift_1$)}

The original target word list for the first pair of time bins was created in \newcite{kutuzov-kuzmenko}. It consists of 43 words that have undergone semantic changes through the period from the pre-Soviet to the Soviet times. The words were collected from general linguistic studies on lexical semantic change \cite{ozheg,twentywords};  there are 38 nouns and 5 adjectives. After the procedure of filler generation, the total number of words for annotation was 71. We emphasize again that only the changed word list was taken from  \newcite{kutuzov-kuzmenko}, the fillers and annotations were recreated from scratch by us. 

\subsubsection{Soviet to Post-Soviet dataset ($RuSemShift_2$)}

The second test set is entirely our own contribution. It also contains manually chosen target words (35 nouns and 7 adjectives), primarily from the  \textit{`New Words and Meanings'} dictionary \cite{dic}. The dictionary includes words which acquired new common senses in Russian in the post-Soviet time period (such words have a special label in the dictionary). Note that these words are not neologisms: all of them occurred in the Soviet sub-corpus as well, so semantic change cannot be estimated by frequencies alone (as shown in Section~\ref{sec:evaluation}). 
After the filler generation procedure, the test set consists of 69 words. 

\section{Annotation} \label{sec:annotation}

The annotation procedure was carried out on the \textit{Yandex.Toloka}  crowd-sourcing platform\footnote{\url{https://toloka.yandex.ru/}}, which is more or less equivalent to the global Mechanical Turk or CrowdFlower platforms. Note that in the DURel dataset \cite{durel}, all five annotators were students of linguistics and two of them had historical background, which cannot be enforced when using crowd-sourcing platforms. However, \textit{Yandex.Toloka} allowed us to improve the quality of annotation by applying  various filters to limit who can annotate. We chose 10-30\% of the best annotators across the whole platform, keeping only native speakers of Russian, of the age 30 and more (to ensure them being familiar with older word senses) and possessing a university degree. We do not have any knowledge about the annotators' gender distribution.

The annotators who completed the tasks unrealistically fast were automatically filtered out. Furthermore, there were several \textit{control tasks}, consisting of sentence pairs manually annotated by ourselves (with the scores of 1 and 4 only, thus limited to obvious cases). The users who annotated such pairs incorrectly were blocked as well. Also, we accepted the judgments of an annotator only after checking some of them manually to make sure that the annotator had understood the task.

\begin{table}
\centering
\begin{tabular}{l l l}
\toprule
\textbf{Value} & \textbf{English judgments from \newcite{durel} } & \textbf{Judgments translated into Russian} \\
\midrule
0 & Cannot decide & \foreignlanguage{russian}{Не знаю}\\
1 & Unrelated & \foreignlanguage{russian}{Значения разные}\\
2 & Distantly Related & \foreignlanguage{russian}{Значения отдаленно похожи}\\
3 & Closely Related & \foreignlanguage{russian}{Очень похожие значения}\\
4 & Identical & \foreignlanguage{russian}{Употреблено в одном значении}\\
\bottomrule
\end{tabular}
\caption{Original 4-point scale from the DURel and its Russian translation used for the annotation.}
\label{tab:4pointscale}
\end{table}

\begin{figure}[th]
\centering
\includegraphics[scale=0.5,keepaspectratio]{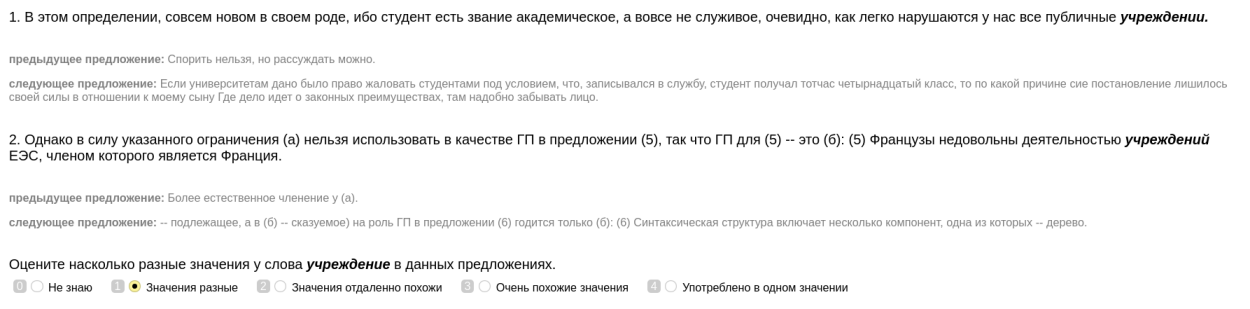}

\caption{Interface of the annotation task. In the first sentence, the word \foreignlanguage{russian}{`учреждение'} [uchrezhdenie] is used in the sense of \sense{decree, order}. In the second sentence, \foreignlanguage{russian}{`учреждение'} denotes \sense{institution}.}
\label{fig:Toloka}
\end{figure}

For each word from our pre-constructed word lists, we extracted all unique sentence contexts from two corpora (\textit{pre-Soviet/Soviet} or \textit{Soviet/post-Soviet}) and randomly sampled 60 sentence pairs, 20 for each group (EARLIER, LATER and COMPARE). For the words that occurred less than 30 times in one of the corpora  we decreased the number of its sentence pairs to the closest value, divisible by 3 without a remainder. Final tables for the annotation contained $7846$ pairs for $71 + 69 = 140$ words. Each sentence pair was judged by not less than 5 annotators who used the scale presented in Table~\ref{tab:4pointscale}. The annotator interface of \textit{Yandex.Toloka} is shown in Figure~\ref{fig:Toloka}. For the cases when one sentence context was not enough to understand the meaning of the target word, we implemented an option to show extended contexts from the same corpora. Annotators were provided with detailed guidelines explaining the task and giving examples for each grade. 

After all sentence pairs were annotated, we measured inter-rater agreement as Krippendorff’s $\alpha$ \cite{alpha} with ordinal scale, excluding 0 judgements (`cannot decide'). For $RuSemShift_1$, the agreement coefficient was 0.505, and for $RuSemShift_2$ its value was 0.53. Considering that the task is inherently ambiguous and complex, and also that all five annotators were different for each sentence pair, we believe this score is high enough. However, we also provide the filtered versions of both test sets, where controversial words that were annotated inconsistently (inter-rater agreement less than 0.2) are excluded. 24 words were filtered out from the $RuSemShift_1$ and 19 from the $RuSemShift_2$). We use these filtered test sets for evaluation in Section~\ref{sec:evaluation}.

\subsection{Analysis}

\begin{wrapfigure}{i}{0.6\textwidth}
\centering
\includegraphics[scale=0.3,keepaspectratio]{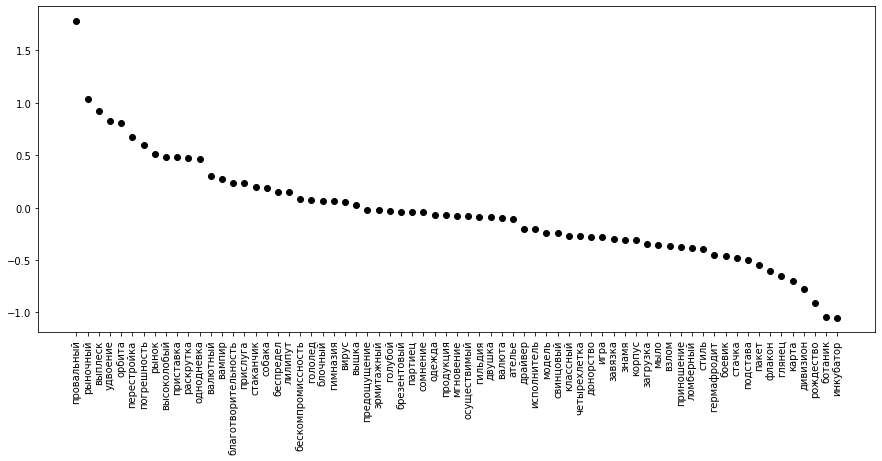}
\caption{Target words ($RuSemShift_2$) ranked by their $\Delta$LATER values.}
\label{fig:Delta}
\end{wrapfigure}
As stated in \newcite{durel}, both measures have their inherent limitations. $\Delta$LATER can fail to capture semantic change if a word loses an old sense and gains a new one within one time period. COMPARE tends to mix up polysemy with semantic change because of random choice of samples (can arguably be remedied using some kind of normalization). $\Delta$LATER naturally captures the differences between two types of meaning change: innovative shift (negative values) or reductive shift (positive values). But this is true only for high absolute $\Delta$LATER values. \newcite{durel} claim COMPARE to be more suitable for indicating the degree of semantic change, but prospective \textit{RuSemShift} users can choose any of these two measures or implement their own (since the raw annotation data is available).

\begin{figure}
\begin{tabular}{ll}
\includegraphics[scale=0.5]{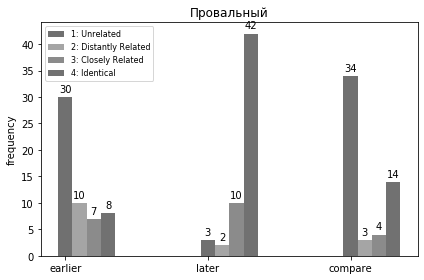}
&
\includegraphics[scale=0.5]{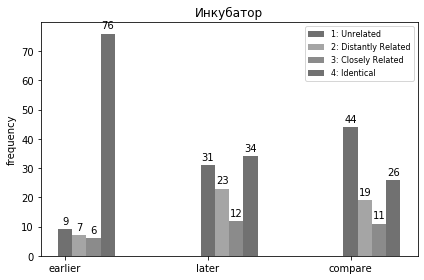}
\end{tabular}
\caption{Distribution of human judgements for the $RuSemShift_2$ target words \foreignlanguage{russian}{`провальный'} `failed' (reductive change) and \foreignlanguage{russian}{`инкубатор'}  `incubator' (innovative change).}
\label{fig:Distr}
\end{figure}

Figure~\ref{fig:Delta} shows the words from the $RuSemShift_2$ ranked according to their $\Delta$LATER values. For most words, it is close to $0$ (no change), but one can also see groups with $\Delta LATER >> 0 $ and $\Delta LATER << 0$, indicating changes. Figure~\ref{fig:Distr} compares the distribution of annotators' judgments for two words with the highest and the lowest $\Delta$LATER values  (the adjective \foreignlanguage{russian}{`провальный'} [proval'nyj] `failed' and the noun \foreignlanguage{russian}{`инкубатор'} [inkubator] `incubator' correspondingly).

For \foreignlanguage{russian}{`провальный'} (the $\Delta$LATER value of $1.78$), we can see that judgements are quite diverse in the EARLIER group and the 1-judgement (`unrelated senses') is prevalent, but in the LATER group, the 4-judgement (`identical senses') is considerably more frequent, while the COMPARE group also captures strong change, since the number of 1-judgements is higher than the number of 4-judgements. Almost all context pairs from the EARLIER group (that is, the Soviet time period) shown in Table~\ref{tab:proval_context} are related to the literal meaning of \foreignlanguage{russian}{`провал'}: \sense{a place where the surface collapsed inward} or the figurative sense of \sense{loss of consciousness} used frequently in the set expression \foreignlanguage{russian}{`провальный сон'} \textit{(`deep dream')}. In all the contexts from the LATER group (the post-Soviet time period), \foreignlanguage{russian}{`провальный'} is used in the sense of \sense{failed} which is more common in modern Russian. Thus, we can observe the expansion of this sense. As for the COMPARE group, sentences in pairs from each period support the same observation: there are only two usages from the earlier period that can be interpreted with the \sense{failed} sense. The word \foreignlanguage{russian}{`провальный'} did not lose its literal meaning, it just became much less frequent, and the old figurative meaning (as in \textit{`deep dream'}) is almost completely lost.\footnote{But is kept in the related word \foreignlanguage{russian}{`провал'} (\foreignlanguage{russian}{`провал в памяти'}/\textit{`lapse of memory'}).} Consequently, we can observe the word losing an old sense and gaining a new one at the same time.

\begin{table}
\small
\centering
\begin{tabular}{ l | p{6cm}| p{6cm} }
\toprule
\textbf{Group} & \textbf{Sentence 1} & \textbf{Sentence 2} \\
\midrule
EARLIER & \foreignlanguage{russian}{<...> выпили "микстуры" и спят где-нибудь сном провальным, пьяным <...>} \break \textit{‘<...> they drank a "potion" and sleeping somewhere in \textbf{deep} drunken \textbf{sleep} <...>’} \break \footnotesize{[Viktor Astaf'ev. Pechal'nyj detektiv (1982-1985)]} & \foreignlanguage{russian}{<...> Кщаре доходит до семидесяти пяти метров и что вообще здесь много провальных озер <...>} \break \textit{‘<...> Kshchare reaches seventy-five meters and in general there are many \textbf{deep lakes} <...>’} \break \footnotesize{[V. A. Solouhin. Vladimirskie prosyolki (1956-1957)]} \\
\midrule
LATER & 
\foreignlanguage{russian}{<...> самые "бюджетные" программы оказываются самыми провальными в исполнении <...>} \break \textit{‘<...> the cheapest programs turn out to be the most \textbf{disastrous}<...>’} \break \footnotesize{[Ivan Golikov. Dohodnoe mesto // «Vsluh o…», 2003.05.19]}
& \foreignlanguage{russian}{<...> Подготовка к референдуму в Чечне до прошлой недели носила провальный характер <...>}  \break \textit{‘<...>Preparation for referendum in Chechnya until last week was \textbf{disastrous}<...>’} \break \footnotesize{[Aleksej Makarkin. Krizisnoe upravlenie «chechenizaciej» // «POLITKOM.RU», 2003.03.03]} \\
\midrule
COMPARE & 
\foreignlanguage{russian}{<...> Наденька на минутку забылась провальным сном и когда открыла глаза <...>} \break \textit{‘<..> Nadenka \textbf{fainted} for a minute and when she opened her eyes <..>’} \break \footnotesize{[B. S. ZHitkov. Viktor Vavich. Kniga vtoraya (1941)]} & 
\foreignlanguage{russian}{Провальное выступление сборной России на Олимпиаде <...>} \break \textit{‘The \textbf{failed} performance of the Russian team at the Olympics <...>’} \break \footnotesize 
{[Sergej Podushkin. Opasnye igry na al'pijskom vozduhe. Gornolyzhniki nachinayut sezon // «Izvestiya», 2002.10.25]} \\ \midrule
\end{tabular}
\caption{Examples of sampled sentence pairs for the word \foreignlanguage{russian}{`провальный'} `failed'.}
\label{tab:proval_context}
\end{table}

For the word \foreignlanguage{russian}{`инкубатор'} (the $\Delta$LATER value of $-1.05$), the distributions are the opposite. The 4-judgment is prevalent in the EARLIER group and there is diversity in the LATER group. Indeed, in the EARLIER group (the Soviet period), \foreignlanguage{russian}{`инкубатор'} is used mostly in its literal meaning of \sense{incubator}, while in the LATER group (he post-Soviet period) there are many occurrences in the figurative sense of \sense{business incubator}. Thereby, the word \foreignlanguage{russian}{`инкубатор'} is undergoing an innovative meaning change in the post-Soviet times, as its figurative sense is becoming more and more widespread.

\section{Evaluation} \label{sec:evaluation}

\textit{RuSemShift} is mainly intended to be used by other researchers in the field of lexical semantic change detection for Russian. However, below we report the performance of several well-known change detection methods on our datasets, to set the baseline. We solved the task of ranking: that is, given a list of target words and two time-specific corpora, a system should predict semantic change degrees which would position the target words in the order as close to the gold one as possible. First we employed static distributional embeddings trained with the CBOW algorithm \cite{Mikolov_representation:2013}. After that, we tried several variations of contextualized embeddings, namely ELMo \cite{elmo}.

\subsection{Static embeddings}

In Table~\ref{tab:baseline}, we report the performance of the method based on static embeddings. We also use a very simple frequency-based baseline, where the degree of semantic change is estimated by the difference of target word absolute frequencies between two time periods. The `word2vec with Procrustes alignment' is the classic method of calculating cosine similarity between target word vectors in two CBOW embedding models trained on different time periods (we used the same splits of the Russian National Corpus as during the creation of \textit{RuSemShift}). The trained models were aligned using Orthogonal Procrustes as described in \newcite{hamilton2016cultural}.

We report the correlations between the predictions of these methods and the COMPARE and $\Delta$LATER values from $RuSemShift_1$ and $RuSemShift_2$. Note that we used the absolute values of $\Delta$LATER, leaving the distinguishing of innovative versus reductive semantic change for future work. This means that an ideal system will produce perfectly positive correlations with $\Delta$LATER and perfectly negative correlations with COMPARE (or vice versa, depending on the exact method), since the former increases as the degree of semantic change grows, while the latter increases as word usages become more similar.

\begin{table}
    \centering
    \begin{tabular}{l|cc|cc}
         \toprule
         \textbf{Dataset} & \multicolumn{2}{c|}{\textbf{$\Delta$LATER}} & \multicolumn{2}{c}{\textbf{COMPARE}} \\
         \midrule
         & Frequency & word2vec Procrustes & Frequency & word2vec Procrustes  \\
         \midrule
         $RuSemShift_1$ & -0.275 & \textbf{0.234} & 0.046  & \textbf{0.403*}  \\
         $RuSemShift_2$ & -0.024 & \textbf{0.118}  & -0.062  & \textbf{0.269}  \\ 
         \bottomrule
    \end{tabular}
    \caption{Spearman $\rho$ correlations of the frequency-based and word2vec-based predictions with \textit{RuSemShift} annotations. * denotes statistical significance at $p<0.05$.}
    \label{tab:baseline}
\end{table}

As expected, comparing cosine distance of Procrustes-aligned word embeddings far outperforms simply measuring frequency changes (this latter method produces predictions close to random in most cases). It can also be observed that $RuSemShift_2$ is more difficult than $RuSemShift_1$. A possible explanation is that Soviet and post-Soviet texts are on average less distant in time from each other than Soviet and pre-Soviet texts: the former pair of time bins lies entirely within 100 years, while the latter covers the time span of about 250 years. Because of that, semantic differences from $RuSemShift_1$ (between Soviet and pre-Soviet lexical meanings) are manifested more clearly in the corpora. Finally, $\Delta$LATER rankings are more difficult to reproduce than those for COMPARE; no method managed to achieve a statistically significant correlation in this case. This arguably stems from the nature of this measure: even though we used its absolute values, it is still focused rather on the nature of semantic shifts than on their degree, and this is not something which can be easily approximated by cosine similarity between Procrustes-aligned word embeddings. More advanced methods are required to better predict $\Delta$LATER values from data.

\subsection{Contextualized embeddings}
We trained ELMo \cite{elmo} models\footnote{Using the implementation from \url{https://github.com/ltgoslo/simple_elmo} with 3 epochs and the vocabulary of 100 000 most frequent words. The rest hyperparameters were left at their default values.} on the RNC texts to produce contextualized token representations for each time period. ELMo embeddings are inferred from bidirectional language models trained using two-layer long short-term memory network (LSTM) for next word predictions in both directions. Every token is represented as a linear combination of hidden layers and depends on the context in which the token appears.

All corpora were segmented into sentences, tokenized and lemmatized with a \textit{UDPipe 1.2} \cite{udpipe} model trained on the SynTagRus treebank \cite{ud_rus}. It is not yet well known whether contextualized embedding models should be trained on lemmatized or non-lemmatized texts, but \newcite{kutuzov-kuzmenko-lemma} showed that at least for Russian, lemmatization improves the performance in word sense disambiguation task. It also excludes word form bias, since we want to trace semantic shifts in lexemes rather than specific word forms.

We trained six ELMo models in three variants:
\begin{enumerate}
    \item a single model trained on the full RNC corpus with texts from all time periods (differentiation by time periods is made at the inference stage when token embeddings are produced);
    \item three models trained separately on each sub-corpus: pre-Soviet, Soviet and post-Soviet models;
    \item two models trained incrementally (initialized from the checkpoint of the model trained on texts from the previous time period): Soviet incremental model and post-Soviet incremental model.
\end{enumerate}

We extract ELMo  token embeddings for each word's usage in two adjacent time periods and estimate semantic change score for this word using the measures described below. Extracted contextualized embeddings of each target word from two time periods are represented as two time-specific matrices. We explored two semantic change detection measures: \textit{cosine similarity between averaged token embeddings}\footnote{This was not used with separately trained models, because it does not make sense to directly measure it between vectors from two different spaces.} and \textit{Jensen-Shannon divergence} which requires prior application of the clustering algorithm.

\begin{enumerate}
  \item \textit{Cosine similarity}. We compute average vectors from usage matrices, which gives us representations which resemble static type embeddings. Then we compute cosine similarity between these average embeddings as a measure of semantic change. The lower is the cosine similarity, the higher is the degree of semantic change.  This method is inspired by the PRT technique from \newcite{kutuzov2020uio}.
  
  \item \textit{Jensen-Shannon divergence} (JSD). In this measure, influenced by \newcite{dubossarsky2015bottom}, \newcite{addmore_2020} and \newcite{giulianelli2020}, word usage matrices from two time periods are first stacked into one matrix. Then, we standardize the vectors and obtain word usage clusters of token embeddings using the Affinity Propagation clustering algorithm \cite{frey2007affinity}. After obtaining clusters for each word, we calculate usage type (sense) probability distributions  for each time period by normalizing counts of word usages in the clusters. Then we compute the JSD score: 
  \begin{equation}
JSD = \sqrt{\frac{D(p \parallel m) + D(q \parallel m)}{2}}
\end{equation} 
\\ where \emph{D} is the Kullback-Leibler divergence, \emph{p} and \emph{q} are sense distributions and \emph{m} is the pointwise mean of \emph{p} and \emph{q}. 
Higher JSD score indicates more intense change in the proportions of clustered word usage types across time periods.
\end{enumerate}

\begin{table}[th]
\small
\centering
\hspace*{-1cm}
\begin{tabular}{c|c|c|c|c|c}
\toprule
\textbf{Model type}  &\textbf{Dataset} & \multicolumn{2}{c|}{\textbf{Cosine similarity}} & \multicolumn{2}{c}{\textbf{JSD}} \\
 \midrule
&  & \multicolumn{1}{l}{\textbf{$\Delta$LATER}} & \textbf{COMPARE} & \multicolumn{1}{l}{\textbf{$\Delta$LATER}} & \textbf{COMPARE} \\
\midrule
\multirow{2}{*}{Single} & $RuSemShift_1$ & \textbf{-0.351}* & 0.246 & -0.158 & \textbf{0.385}* \\
 & $RuSemShift_2$ & \textbf{-0.300}* & \textbf{0.541}* & -0.147 & \textbf{0.364}* \\
\midrule
\multirow{2}{*}{Incremental} & $RuSemShift_1$ & 0.068 & 0.055 & -0.117 & -0.074 \\
 & $RuSemShift_2$ & 0.028 & 0.186          & 0.028 & 0.186 \\
\midrule
\multirow{2}{*}{Separate} & $RuSemShift_1$ & - & - & 0.201 & -0.283 \\
 & $RuSemShift_2$ & - & - & -0.064 & 0.100    
\end{tabular}
\caption{Spearman $\rho$ correlations of the ELMo-based predictions with \textit{RuSemShift} annotations.}     
\label{tab:scores}
\end{table}

Table~\ref{tab:scores} shows that incremental and separate ELMo models do not yield significant correlations with human judgments. As for the single model, it consistently outperforms static embeddings on $RuSemShift_2$ (both measures), and $RuSemShift_1$ ($\Delta$LATER): thus, in 3 out of 4 cases.
As already observed, COMPARE metrics is easier to approximate than $\Delta$LATER. Also, cosine similarity generally is better than JSD, despite the latter being much heavier computationally. Negative correlations with $\Delta$LATER are normal and caused by the nature of this measure: higher values indicate stronger change. 

\section{Conclusion}
\label{conclusion}

We presented \textit{RuSemShift}, which consists of two publicly available test sets of Russian words manually annotated with the degrees of diachronic semantic change they experienced (the degree is a continuous score). Annotation process was based on the theoretically sound DURel framework \cite{durel}. One of the datasets is produced from re-annotated list of target words from prior work and another is completely new. The datasets allow to evaluate methods for lexical semantic change detection in Russian: either graded or binary (with any desired binarization threshold). They provide data on 3 large time periods: pre-Soviet (1682-1916), Soviet (1918-1990) and post-Soviet (1991-2017).

This is the first semantic change detection dataset for Russian created in a large-scale crowd-sourcing annotation effort. It is also important that \textit{RuSemShift} is fully compatible with semantic change datasets developed for other languages, for example those presented for the corresponding SemEval-2020 shared task \cite{schlechtweg2020semeval}. The dataset is available online\footnote{\url{https://github.com/juliarodina/RuSemShift}} under a Creative Commons Attribution-ShareAlike 4.0 International License.

As a sanity check (and to establish the baseline performance boundaries), we evaluated several semantic change modeling systems on  \textit{RuSemShift}. We managed to achieve significant correlation with human judgments both with static and with contextualized word embeddings, with the latter consistently outperforming the former. At the same time, simple frequency-based baseline failed to achieve any meaningful results, which signals that the dataset lacks simplistic frequency cues.

\textit{RuSemShift} is limited to nouns and a few adjectives. One of the possible future research directions is to extend the dataset with other parts of speech. Another drawback of the dataset is that it does not distinguish between different types of semantic change (e.g. narrowing, widening, metaphorization etc.). Nevertheless, we hope that in its current state, \textit{RuSemShift} will already  be of help to the researchers interested in tracing diachronic semantic shifts in Russian.

\section*{Acknowledgments}
This research was supported by the Russian Science Foundation grant 20-18-00206.

\bibliographystyle{coling}
\bibliography{coling2020}

\end{document}